\documentclass[conference]{IEEEtran}
\IEEEoverridecommandlockouts
\usepackage[margin=0.7in]{geometry}
\usepackage{cite}
\usepackage{amsmath,amssymb,amsfonts}
\usepackage{graphicx, caption, subcaption}
\usepackage{textcomp}
\usepackage{dblfloatfix}
\usepackage[table]{xcolor}
\usepackage{algorithm,algpseudocode}
\usepackage{multirow}
\usepackage{hyperref}
\hypersetup{
     colorlinks = true,
     citecolor = green,
     urlcolor = blue
     }

\usepackage{multirow}
\usepackage{booktabs}
\usepackage[usestackEOL]{stackengine}
\usepackage{lipsum}
\def\BibTeX{{\rm B\kern-.05em{\sc i\kern-.025em b}\kern-.08em
    T\kern-.1667em\lower.7ex\hbox{E}\kern-.125emX}}
\begin{document}


\newboolean{showcomments}
\setboolean{showcomments}{true} 
\ifthenelse{\boolean{showcomments}}
  {
		\newcommand{\nbb}[2]{
		\fcolorbox{black}{yellow}{\bfseries\sffamily\scriptsize#1}
		{\sf$\blacktriangleright$\textcolor{blue}{\textit{#2}}$\blacktriangleleft$}
		}
		\newcommand{\version}{\emph{\scriptsize$-$9.2.2011$-$}}
		\newcommand{\remarks}[1]{\color{red}[#1]\color{black}}
		\newcommand{\copied}[1]{\color{green}[#1]\color{black}}
		\newcommand{\modified}[1]{\color{blue}[#1]\color{black}}
		\newcommand{\raw}{$\rightarrow$}
		\newcommand{\ins}[1]{\textcolor{blue}{\uline{#1}}} 
		\newcommand{\del}[1]{\textcolor{red}{\sout{#1}}} 
		\newcommand{\chg}[2]{\textcolor{red}{\sout{#1}}{\raw}\textcolor{blue}{\uline{#2}}} 
		\newcommand{\ugh}[1]{\textcolor{red}{\uwave{#1}}} 
  }
  {
		\newcommand{\nbb}[2]{}
		\newcommand{\remarks}[1]{}
		\newcommand{\modified}[1]{#1}
		\newcommand{\copied}[1]{#1}
		\newcommand{\version}{}
		\newcommand{\ugh}[1]{#1} 
		\newcommand{\ins}[1]{#1} 
		\newcommand{\del}[1]{} 
		\newcommand{\chg}[2]{#2} 
  }

\newcommand{\jens}[1]{\nbb{Jens}{#1}}
\newcommand{\cbe}[1]{\nbb{CBe}{#1}}
\newcommand{\lars}[1]{\nbb{Lars}{#1}}
\newcommand{\sankar}[1]{\nbb{Sankar}{#1}}
\newcommand{\ce}[1]{\nbb{CE}{#1}}
\newcommand{\mb}[1]{\nbb{Markus}{#1}}
\newcommand{\comment}[1]{\nbb{Comment}{#1}}

\title{Performance Analysis of Out-of-Distribution Detection on Trained Neural Networks}

\author{
\IEEEauthorblockN{Jens Henriksson\IEEEauthorrefmark{1}, Christian Berger\IEEEauthorrefmark{2}, Markus Borg\IEEEauthorrefmark{3},\\ Lars Tornberg\IEEEauthorrefmark{4}, Sankar Raman Sathyamoorthy\IEEEauthorrefmark{5}, Cristofer Englund\IEEEauthorrefmark{3}}\\
\IEEEauthorblockA{\IEEEauthorrefmark{1}Semcon AB, Gothenburg, Sweden, Email: jens.henriksson@semcon.com}
\IEEEauthorblockA{\IEEEauthorrefmark{2}University of Gothenburg and Chalmers Institute of Technology, Sweden, Email: christian.berger@gu.se}
\IEEEauthorblockA{\IEEEauthorrefmark{3}RISE Research Institutes of Sweden AB, Lund and Gothenburg, Sweden, Email: \{markus, cristofer\}@ri.se}
\IEEEauthorblockA{\IEEEauthorrefmark{4}Machine Learning and AI Center of Excellence, Volvo Cars, Gothenburg, Sweden, Email: lars.tornberg@volvocars.com}
\IEEEauthorblockA{\IEEEauthorrefmark{5}QRTech AB, Gothenburg, Sweden, Email: sankar.sathyamoorthy@qrtech.se}}

\maketitle

\begin{abstract}

Several areas have been improved with Deep Learning during the past years. Implementing Deep Neural Networks (DNN) for non-safety related applications have shown remarkable achievements over the past years; however, for using DNNs in safety critical applications, we are missing approaches for verifying the robustness of such models. A common challenge for DNNs occurs when exposed to out-of-distribution samples that are outside of the scope of a DNN, but which result in high confidence outputs despite no prior knowledge of such input.

In this paper, we analyze three methods that separate between in- and out-of-distribution data, called supervisors, on four well-known DNN architectures. We find that the outlier detection performance improves with the quality of the model. We also analyse the performance of the particular supervisors during the training procedure by applying the supervisor at a predefined interval to investigate its performance as the training proceeds. We observe that understanding the relationship between training results and supervisor performance is crucial to improve the model's robustness and to indicate, what input samples require further measures to improve the robustness of a DNN. In addition, our work paves the road towards an instrument for safety argumentation for safety critical applications. This paper is an extended version of our previous work presented at 2019 SEAA (cf.~\cite{henriksson2019performance}); here, we elaborate on the used metrics, add an additional supervisor and test them on two additional datasets.

\end{abstract}


\section{Introduction}
Deep Neural Networks (DNN) constitute a major challenge to safety engineers, especially for complex and safety-critical functionality such as Autonomous Driving. Autonomous Driving requires complex perception systems embodying DNNs for computer vision to process large streams of video, lidar, and radar sensors. Even though DNNs have revolutionized the field of computer vision~\cite{lecun_deep_2015}, they still pose an enormous challenge for safety verification and validation since their characteristics and development processes are completely different compared to conventional software addressed by the automotive safety standard ISO~26262 -- Functional Safety~\cite{international_organization_for_standardization_iso_2018}. The behaviour of a DNN is not explicitly expressed by an engineer in source code following the principle, where the developer defines an algorithm based on a specification, but rather instead the developer defines an architecture that learns its connections. To achieve this learning, developers usually require large amounts of data complemented with domain-specific labels to train specific behavior. For autonomous driving, this step usually comprises of data collection, preprocessing/cleaning, training, and evaluation of large quantities of annotated camera, lidar, and radar data using Machine Learning (ML)~\cite{salay_analysis_2017,henriksson_automotive_2018}.

\subsection{Background}

In January 2019, ISO/PAS~21448 -- Safety of the Intended Functionality (SOTIF) was published to meet a growing industrial need of an automotive safety standard that is appropriate for ML~\cite{international_organization_for_standardization_iso_2019}. A PAS (Publicly Available Specification) is not an established standard, but a document that closely resembles what is planned to become a future standard; a PAS is published to speed up a standardization process in response to market needs. A PAS must be transformed to a standard within six years, otherwise it will be withdrawn\footnote{https://www.iso.org/deliverables-all.html}.

ISO~26262 and SOTIF are meant to be complementary standards: ISO~26262 covers ``absence of unreasonable risk due to hazards caused by malfunctioning behaviour'' \cite{international_organization_for_standardization_iso_2018} by mandating risk reduction activities that is achieved by following the established V-model. SOTIF, on the other hand, addresses ``hazards resulting from functional insufficiencies of the intended functionality''~\cite{international_organization_for_standardization_iso_2019}, e.g., miss-classifications by an object detector in an automotive perception system. To achieve systematic risk reduction of functional insufficiencies, ML requires quantitative requirements in addition to the qualitative instruments that are currently present in ISO~26262.

SOTIF is not organized according to the V-model but instead around four states defining all possible events: 1) known safe states, 2) known unsafe states, 3) unknown unsafe states, and 4) unknown safe states. SOTIF presents a process to minimize the two unsafe states, represented by a flowchart orbiting around requirements specifications for the functionality under development, where moving hazards from 3) $\rightarrow$ 2) and from 2) $\rightarrow$ 1) originates from hazard identification and hazard mitigation, respectively.  

\subsection{Problem Domain \& Motivation}

Following the SOTIF process, we identify that autonomous driving in environments that differ too much from what the automotive perception was trained for, constitutes a major hazard; in ML literature, this is referred to as dealing with out-of-distribution (OOD) samples. In previous work, we have proposed various DNN supervisors~\cite{henriksson_automotive_2019} to complement autonomous driving by detecting such outlier samples as part of a safety cage architecture~\cite{borg_safely_2019}, i.e., adding a reject option to DNN classifiers when input does not resemble the training data.

As part of a mitigation strategy for out-of-distribution camera data for example, we argue that the SOTIF process should result in a safety requirement mandating the inclusion of a DNN supervisor. However, only a few previous studies have presented structured evaluations of different DNN supervisors for image data; most previous work instead focuses on a single supervisor, like, for example, \cite{bendale2016towards,liang2017enhancing} and \cite{hendrycks2018deep}.

\subsection{Research Goal \& Research Questions}

In this paper we evaluate three supervisor algorithms: a baseline algorithm~\cite{hendrycks2016baseline},  OpenMax~\cite{bendale2016towards} and ODIN~\cite{liang2017enhancing}. In this context, their algorithms have been adjusted to act as a supervisor and provide an anomaly score of each input sent to the model to comply with the test setup defined in \cite{henriksson_automotive_2019}. Our evaluation method consists of applying three supervisors on four established DNNs architectures, namely VGG16~\cite{DBLP:journals/corr/SimonyanZ14a}, DenseNet-121~\cite{DBLP:journals/corr/HuangLW16a} and two versions of Wide ResNets~\cite{DBLP:journals/corr/ZagoruykoK16}. All networks are trained with identical hyperparameters on the CIFAR-10 dataset~\cite{krizhevsky2009learning} and tested on three different outlier sets. We address the following two research questions:

\begin{itemize}
    \item How does the outlier detection performance change with improved training of the DNNs under supervision?
    \item How does overfitting of the DNNs under supervision affect the supervisor performance?
\end{itemize}


\hfill 

\subsection{Contributions}

We find that the performance of the supervisor is almost linear with regards to the network performance. However, the design and tuning of supervisors can be a sensitive development process, especially when applied at different stages during training where network parameters have been shifted. This paper is an extended version of our previous work presented at 2019 SEAA (cf.~\cite{henriksson2019performance}); here, we elaborate on which metrics we use, add an additional supervisor to the comparison, as well as extend the testing to one additional image classification dataset and one dataset made out of noise. 

\subsection{Structure of the Article}

The rest of the paper is organized as follows: Section~\ref{sec:rw} introduces related work on automotive perception and DNN supervisors. In Section~\ref{sec:eval}, we describe our method to evaluate and compare DNN supervisors. Section~\ref{sec:results} presents our results and in Section~\ref{sec:disc} we discuss our findings in the light of safety for autonomous driving. Finally, Section~\ref{sec:conc} concludes our paper and outlines directions for future work.

\section{Related Work}
\label{sec:rw}

Recent work on AI/ML has gained a lot of attention for the break-through of large-scale pattern recognition for various applications ranging from popular old board games, over text and image analysis, up to complete end-to-end solutions for autonomous driving. Even though the reported results appear fascinating and underline the potential of AI/ML, the challenges of commercializing such solutions into safe products is the apparent next challenge to overcome for a successful roll-out. Hence, recent works are now exploring and discussing how to test deep learning models. Fei et al.~\cite{deepXplore} and the succeeding study from Guo et al.~\cite{DLFuzz} suggest to evaluate testing coverage of models by evaluating neuron coverage. While such approaches show similarities to code, statement, or branch coverage in software testing as defined in ISO~26262~\cite{international_organization_for_standardization_iso_2018}, they fall short for evaluating the robustness of DNNs as increased neuron coverage as achieved by artificially constructed test samples does not necessarily correlate with out-of-distribution samples that a DNN would experience in reality. Furthermore, the improvements of these approaches reported by the authors are in the range of 2-3\%. Kim et al.~\cite{GuidedDLTesting} suggested a different approach, where they analyzed the influence of a particular stimulus to a DNN with the purpose of providing support when designing specific test data to evaluate the performance of a DNN.

Czarnecki and Salay postulate in their position paper \cite{Czarnecki_2018} a framework to manage uncertainty originating from perception-related components with the goal of providing a performance metric. Their work provides initial thoughts and concepts of how to describe uncertainty that is present in ML-based approaches. However, in contrast to our work here, uncertainty originating from the trained model itself is only captured briefly and not in a quantitative approach as outlined here.  

A different approach is presented by Ma, Driggs-Campbell, and Kochenderfer, who investigate in their work the problem of robust learning for autonomous vehicle control by formulating it as a two-players-game between the autonomous system and disturbances \cite{Ma_2018}. They introduce two new concepts to this game theoretic formulation. In Ma et al.~\cite{ma2018deepgauge}, DeepGauge is introduced, which proposes several multi-granularity testing criteria for DL systems aiming at rendering a multi-faceted portrayal of the testbed. Initial results from experiments show that their approach might be applicable to evaluate the testing adequacy of DNNs. 

Several recent papers have discussed the topic of out-of-distribution detection. Hendrycks and Gimpel \cite{hendrycks2016baseline} described an approach to monitor the performance of a DNN-based system by evaluating the Softmax layer. This is used to predict if the network is classifying a given sample correctly; in addition, they also use this feature as a discriminator between in-distribution and out-of-distribution from the training data.  

Similarly to Hendrycks and Gimpel, Liang et al.~\cite{liang2017enhancing} also use the Softmax layer in their approach named ODIN but without the need for any changes in a given model. Bendale et al.~\cite{bendale2016towards} proposed a new model layer OpenMax, which is a replacement for the Softmax layer. This new layer compares the input sample towards an unknown probability element, adapted from meta-recognition of the penultimate layer, thus allowing the layer to learn the behavior of the training set, and correlated classes.  

Carlini and Wagner~\cite{carlini2017towards} showed how adversarial attacks can break defensive distillation \cite{papernot2016distillation}, a recent approach tailored to reduce the success rate of adversarial attacks from 95\% to 0.5\%. Their adversarial attack is tailored to three distance metrics, which are commonly used when generating adversarial samples. 

Finally, in contrast to only monitoring the Softmax layer, Tranheden and Landgren~\cite{MattiasLudwig18} compared and extracted information from five different detection methods to craft their own. Their combined solution suggests to supervise activations on multiple layers as various sub-models, since the behavior varies depending on input, thus would benefit from model or class specific supervisors.


\section{Evaluation Method} \label{sec:eval}
The scope of this work consists of comparing three implementations of supervisors on neural networks, which are particularly trained for image classification. To achieve a substantiated comparison, the supervisors will be tested on four image classification models. Additionally, as the training results may vary on stochastic effects such as weight initialization or random selection of mini batch gradient descent, we believe by testing every $10^{th}$ epoch, small variations in the model weight will show how the performance of the supervisor changes. It is of interest to study this behavioral change as small perturbations both in network quality or input space can drastically change the model output. In this section, the model training, datasets and evaluation metrics are described in detail. 

\textbf{Model training:} For our experiments, four state of the art models, based on three architectural designs are trained from scratch. The first architecture, VGG16~\cite{DBLP:journals/corr/SimonyanZ14a}, is a convolutional neural network (CNN) with all layers connected in series. The second model is the DenseNet-121 architecture as described in~\cite{DBLP:journals/corr/HuangLW16a}, with internal \textit{dense blocks} that refer to sets of convolutional layers, where previous layer outputs are concatenated to further improve informational flow. The last architecture is the Wide Residual Network (WRN) \cite{DBLP:journals/corr/ZagoruykoK16}, who introduced a study of depth and width for residual networks \cite{he_deep_2016}. A residual block consists of a layer that concatenates the feature input such as $x_{l} = x_{l-1} + f(x_{l-1})$ where $f(\cdot)$ refers to a set of layer permutations of the input. For our experiments, two model setups are selected for WRN, namely WRN-28-10 and WRN-40-10, where the depth factor is set to 28 and 40 respectively, and width factor of 10 for both networks. 

All four networks are trained with identical hyperparameters, for 200 epochs with stochastic gradient descent with momentum. The learning rate is initialized at $10^-1$ and decreased by a factor 10 after 100 and 150 epochs. The model weights are stored after every $10^{th}$ epochs, as well as the best model with highest accuracy on the training set, thus yielding 21 versions each  all models. 

\textbf{Datasets:}
During evaluation, as well as in training the CIFAR-10 dataset will be used. For the evaluation phase, the test set is used as in-distribution (i.e., data samples, which are not excluded). For out-of-distribution samples, we use two datasets with natural images, and one dataset consisting of artificially constructed noise. All datasets are described more specific in the following: 

\begin{itemize}
    \item \textbf{Tiny ImageNet:} Tiny ImageNet consists of a selection of images from the ImageNet \cite{ILSVRC15} dataset that is rescaled to $64x64$. The dataset consist of 200 classes. We further downscale the validation images to match CIFAR-10 image dimensions, as well as use the validation part of the dataset that consists of 10,000 images. 
    \item \textbf{SVHN:} The Street View House Numbers (SVHN) \cite{netzer2011reading} dataset contains more than 600,000 labeled images of small cropped digits with the goal of recognizing digits and numbers in natural scene images. The house numbers are obtained through Google Street View imagery. For our dataset, we use the first 10,000 images from the training set. 
    \item \textbf{FakeData:} The Pytorch\footnote{An open source machine learning framework: \href{https://pytorch.org}{https://pytorch.org}} FakeData dataset generates random samples, where each pixel is sampled from an i.i.d Gaussian distribution. We generate 10,000 random RGB images with mean 0.5 and standard deviation 1.  
\end{itemize}

\begin{algorithm}[!tpb] 
\caption{The Anomaly Score extraction from out-of-distribution methods. Each method computes an activation vector $\textbf{v}(\textbf{x})$ for a given input sample $\mathbf{x}$. The activation vector is then normalized into a probability distribution, that is used to compute the anomaly score.}
\label{Alg:AnomalyScore}
\begin{algorithmic}[1]
\Require threshold $\epsilon$
\Require Out-of-distribution method $F(\cdot)$
\State Compute the output vector \textbf{v}(\textbf{x}) for input sample \textbf{x} with the Out-of-distribution method $F(\cdot)$
\State Let $P$ be the Softmax of output vector \textbf{v} for all classes $j = 1, .., N$
\Statex \begin{equation} P(y=j |\textbf{x})= \frac{ e^{\textbf{v}_{j}(\textbf{x})} }{\sum_{i=1}^{N} e^{\textbf{v}_{i}(\textbf{x})}} \end{equation}
\State Let the anomaly score be $A(\textbf{x})$ = 1 - argmax $P(y=j |\textbf{x})$ where $argmax$ is the function argument that maximizes function value.

\State Let discriminator $D\{0, 1\}$ be defined as \begin{equation}
    D = \left\{\begin{matrix}
\begin{matrix}
1 & if A(\textbf{x}) <  \epsilon \\ 
0 & otherwise
\end{matrix}
\end{matrix}\right.
\end{equation}
\end{algorithmic}
\end{algorithm}

\begin{figure*}[!tbh]
\includegraphics[width=\textwidth]{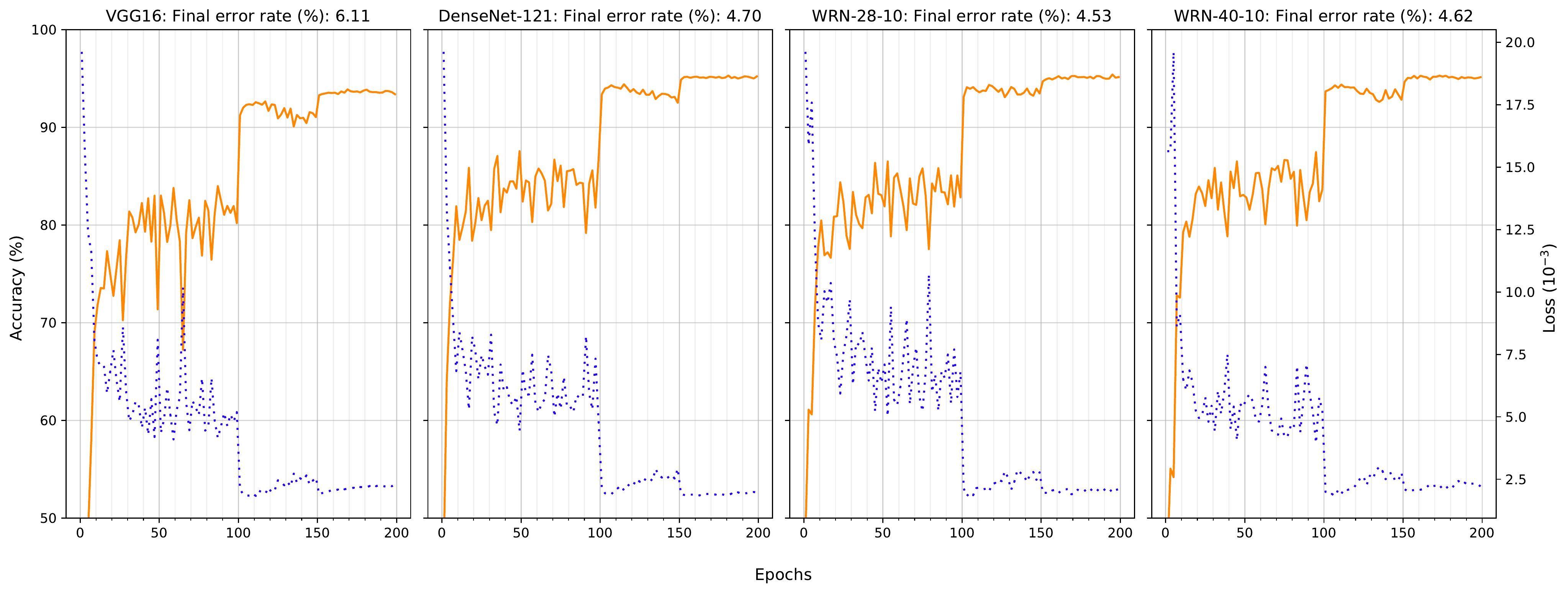}
\caption{The training results for the four models: The orange line represents the testing accuracy achieved during training, and the blue dashed line represents the cross entropy loss. All models are trained with stochastic gradient descent with momentum and $10^{-1}$ learning rate that is reduced by a factor 10 after 100 and 150 epochs.}
\label{fig:training_results}
\end{figure*}

\textbf{Metrics:} Based on recent studies \cite{henriksson_automotive_2019,liang2017enhancing}, we extend on their selection of metrics and select four metrics that show the discriminative rate of the supervisor as well as the overall performance of the network at hand. For our experimental setups, we define a True Positive (TP) as an outlier correctly being selected for removal and a False Positive (FP) as an inlier that has been incorrectly removed. Additionally, in this context True Negatives (TN) refers to inlier samples being accepted and False Negatives (FN) being outliers that has not been rejected. Thus, True Positive Rate (TPR) and False Positive Rate (FPR) can be computed as $TPR=TP/(TP+FN)$ and $FPR=FP/(FP+TN)$. The metrics are explained and motivated below. 

\begin{itemize}
    \item \textbf{AUROC:} Area Under the Receiver Operating Characteristics curve,  which plots the true positive rate (TPR) against the false positive rate (FPR). The curve is received by creating various thresholds in the anomaly score of the supervisor.
    \item \textbf{FPR at $95\%$ TPR (FPR95):} This metric can be interpreted as the probability that an inlier sample will be removed when the majority (95\%) of out-of-distribution samples are rejected. 
    \item \textbf{CBPL:} Coverage break-point at performance level. The metric relates how restrictive the threshold has to be to return to the original accuracy that was received based on test set. 
    \item \textbf{Cov10:} Coverage at $10\%$ Error rate, refers to the coverage break-point but at a given accepted error rate. This metric does not penalize well-scoring models due to requiring a more restrictive threshold.  
\end{itemize}

\textbf{Out-of-distribution methods:} 
This paper studies three methods constructed to detect out-of-distribution samples. The experiments work on the premise of Algorithm \ref{Alg:AnomalyScore}, where each method returns an activation vector that can be used to compute an \textit{anomaly score}. In the algorithm, $F(\cdot)$ refers to the OOD method, and that the threshold $\epsilon$ is varied between the lowest and highest anomaly score to receive the ROC-curve. It is worth noting that some of the methods required tuning, which is done once per method, thus not being optimized per model basis. Additionally, the methods are not built for thresholding the full anomaly score spectra, but rather finding one threshold. Thus, results may look worse due to not taking into account the best threshold per model. The three methods used in this paper are described in the following:

\begin{itemize}
    \item \textbf{Baseline} \cite{hendrycks2016baseline}: Hendrycks et al.~showed baseline performance by looking at the softmax output layer of a neural network. They analyzed several inlier and OOD datasets and showed that by only looking at the most plausible prediction, a majority of OOD samples could be rejected. 
    \item \textbf{ODIN} \cite{liang2017enhancing}: Liang et al.~proposed a method that utilizes the gradients inside a network as well as temperature scaling on the softmax layer to construct a small perturbation on the image. They showed that their perturbation would hurt the inlier samples more than OOD samples, thus utilizing this behavior to differentiate between the distributions. 
    \item \textbf{OpenMax} \cite{bendale2016towards}: Bendale et al.~proposed a method to adjust the softmax layer by incorporating information from the training set. By utilizing meta-recognition algorithms their approach constructs a Weibull-distribution based on the training samples that can be used as a distance metric between the sample in question towards the training set. This information is used to compute a probability of an \textit{unknown unknown} class. Additionally, they are not only doing this revision for the top class, but rather the $n$ top classes, since most inlier samples follow the same order of top classes.
\end{itemize}


\begingroup
\renewcommand*{\arraystretch}{1.3}
\begin{table*}[t]
\centering
\caption{The performance for supervisor on the three Out-of-Distribution datasets, with the best performing (test set accuracy wise) model. All metrics are shown in percentages, where higher indicates better. For each metric, ODIN, OpenMax and Baseline are presented in order. }
\label{tab:best_performance}
\begin{tabular}{llllll}
\hline
 &  & \multicolumn{4}{c}{\textbf{ODIN / OpenMax / Baseline}} \\ \cline{3-6} 
\multicolumn{1}{l}{\textbf{Model}} & \textbf{\begin{tabular}[c]{@{}l@{}}Out-of-distribution\\ dataset\end{tabular}} & \multicolumn{1}{c}{\textbf{AUROC $\uparrow$}} & \multicolumn{1}{c}{\textbf{FPR at 95\% TPR $\downarrow$}} & \multicolumn{1}{c}{\textbf{CBPL $\uparrow$}} & \multicolumn{1}{c}{\textbf{Cov10 $\uparrow$}} \\ \hline
 & \textbf{Tiny ImageNet} & 87.39 / 84.80 / 86.68 & 55.75 / 67.07 / 57.22 & 43.34 / 32.52 / 37.98 & 49.25 / 48.34 / 50.14 \\
\textbf{DenseNet-121} & \textbf{SVHN} & 97.23 / 93.19 / 95.34 & 10.37 / 12.03 / 12.83 & 51.95 / 49.25 / 49.54 & 55.16 / 54.13 / 55.37 \\
 & \textbf{FakeData} & 95.45 / 95.23 / 99.00 & 7.68 / 9.54 / 3.28 & 51.75 / 45.23 / 51.93 & 54.79 / 45.23 / 56.96 \\ \hline
 & \textbf{Tiny ImageNet} & 86.46 / 83.45 / 84.85 & 61.03 / 72.31 / 59.69 & 45.15 / 36.73 / 41.24 & 49.17 / 47.58 / 48.35 \\
\textbf{VGG 16} & \textbf{SVHN} & 88.18 / 85.37 / 89.26 & 45.42 / 66.91 / 31.37 & 47.04 / 41.77 / 46.34 & 50.18 / 49.78 / 51.16 \\
 & \textbf{FakeData} & 5.51 / 5.95 / 32.97 & 99.76 / 99.63 / 97.68 & 8.18 / 7.00 / 8.86 & 10.30 / 10.44 / 13.15 \\ \hline
 & \textbf{Tiny ImageNet} & 89.07 / 87.47 / 87.64 & 48.07 / 51.78 / 46.33 & 43.65 / 39.97 / 41.70 & 50.97 / 50.72 / 50.68 \\
\textbf{WRN-28} & \textbf{SVHN} & 83.50 / 84.38 / 88.82 & 41.59 / 43.28 / 26.74 & 40.96 / 38.68 / 44.70 & 46.97 / 47.38 / 51.05 \\
 & \textbf{FakeData} & 99.77 / 95.56 / 97.71 & 0.54 / 8.47 / 3.95 & 52.90 / 48.14 / 51.77 & 56.66 / 48.14 / 56.51 \\ \hline
 & \textbf{Tiny ImageNet} & 87.79 / 85.39 / 86.48 & 62.87 / 70.66 / 58.10 & 43.28 / 33.99 / 39.48 & 50.58 / 49.23 / 50.32 \\
\textbf{WRN-40} & \textbf{SVHN} & 89.35 / 86.46 / 89.54 & 34.37 / 49.24 / 27.96 & 46.15 / 39.83 / 44.90 & 50.97 / 49.85 / 51.74 \\
 & \textbf{FakeData} & 83.53 / 87.51 / 83.81 & 26.71 / 19.74 / 23.51 & 45.15 / 46.27 / 44.65 & 49.08 / 51.84 / 49.95 \\ \hline
\end{tabular}
\end{table*}
\endgroup

\section{Experimental results} \label{sec:results}
\textbf{Model training:} In Fig. \ref{fig:training_results}, the training process can be seen for each of the four models. The training is conducted with image augmentation that consists normalization, random cropping, and flipping. These augmentations improve generalization of the models by forcing it to learn multiple representations per input. Additionally, batch size was set to 128 and learning rate to $10^{-1}$ that decreased after 100 and 150 epochs by a factor 10.    

The training process contains large variations of accuracy during the first 100 epochs, which is expected due to the high learning rate. Additionally, it benefits the comparison of supervisors since it indicates variations in weights and biases inside the network. As the learning rate decreases, we can see rapid convergence up until a period of overfitting that is occurring roughly in the same interval of epochs for all models. While overfitting is generally not desired for training DNNs, we embrace it as it unravels how it affects the perform ace of supervisors when applied to overtrained models. 

\textbf{Parameter selection:} Both ODIN and OpenMax require parameters for their algorithm to work. As we are interested in a generalized supervisor only one parameter setup exist per model, which is independent of out-of-distribution dataset. This way, one parameter search is per model followed by a selection of the setup that has best AUROC on average. The parameter search is conducted through a grid search. For ODIN, we iterate over the temperature scaling set to 50, 100, 200, 500, 700, 1000, 1500, and 2000; and for the noise gradient we create 21 linear steps between 0.0 and 0.004. For OpenMax, we iterate over tail length set to 5, 10, 20, 50, and 100; and the top alpha classes are set to 1, 2, \dots 10.

\subsection{Results on best epochs} 
In Table \ref{tab:best_performance}, the metric performance for the three supervisors can be compared per OOD-dataset. Regarding the AUROC scores, it is noteworthy that Openmax does not perform better than baseline. This is due to OpenMax being optimized for a single threshold, i.e., optimized to create and use their unknown unknown class score only if it is the most probable outcome. We, however, use it for the full threshold comparison as described in Algorithm \ref{Alg:AnomalyScore}. While most supervisor-model combinations perform similarly on the Tiny ImageNet dataset, the same cannot be observed for SVHN and FakeData. For SVHN, which is another image dataset, the results vary far more than on the Tiny ImageNet but the overall results are better. DenseNet-121 for example, when combined with ODIN, receives a score of 97.23, which indicates a very good separation between inlier and outlier datasets. All models except for VGG16 succeeds to detect samples with random Gaussian noise, whereas VGG seem to completely miss to separate the in and out distribution. In this specific case, the random noise creates larger stimulus than the training samples, and thus, the model believes that outliers are the true distribution. 

\begin{figure*}[!t]
\centering
\includegraphics[width=\textwidth]{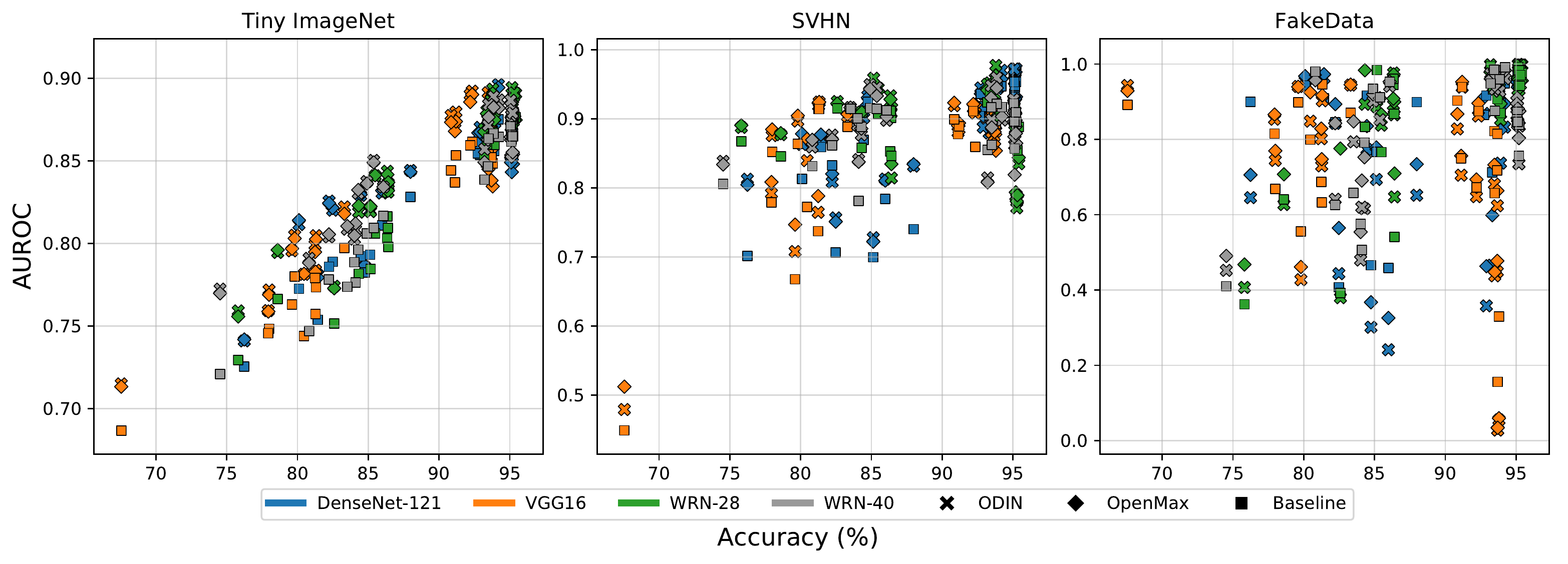}
\caption{The AUROC metric plotted against the test set accuracy for each individual model, on the three out-of-distribution datasets. The coloring represents which model is used, while the marker represent which supervisor is used (i.e., an orange square indicates one of the epochs from VGG16 tested with baseline).}
\label{fig:auroc-vs-accuracy}
\end{figure*}

Similarly to AUROC, the FPR, which we measure when the model removes 95\% of the outliers, shows how rapidly the supervisor can exclude the majority of outliers. On the Tiny ImageNet dataset, it can be seen that most models have high AUROC but also high FPR95. This shows that the supervisor does a good job but the last percentages of outliers cannot be separated. This is explained by the fact that the datasets are not disjoint: Some classes exist in both datasets, which creates a challenge for separation. For SVHN and FakeData, the FPR are performing better, DenseNet-121 does a good job, as well as WRN-28 that does almost perfectly on FakeData but performs poorly on SVHN. This result alone shows how fragile the supervisor performance can be, where performance on one dataset can be high, while failing on another. 

\begin{figure*}[!b]
\includegraphics[width=\textwidth]{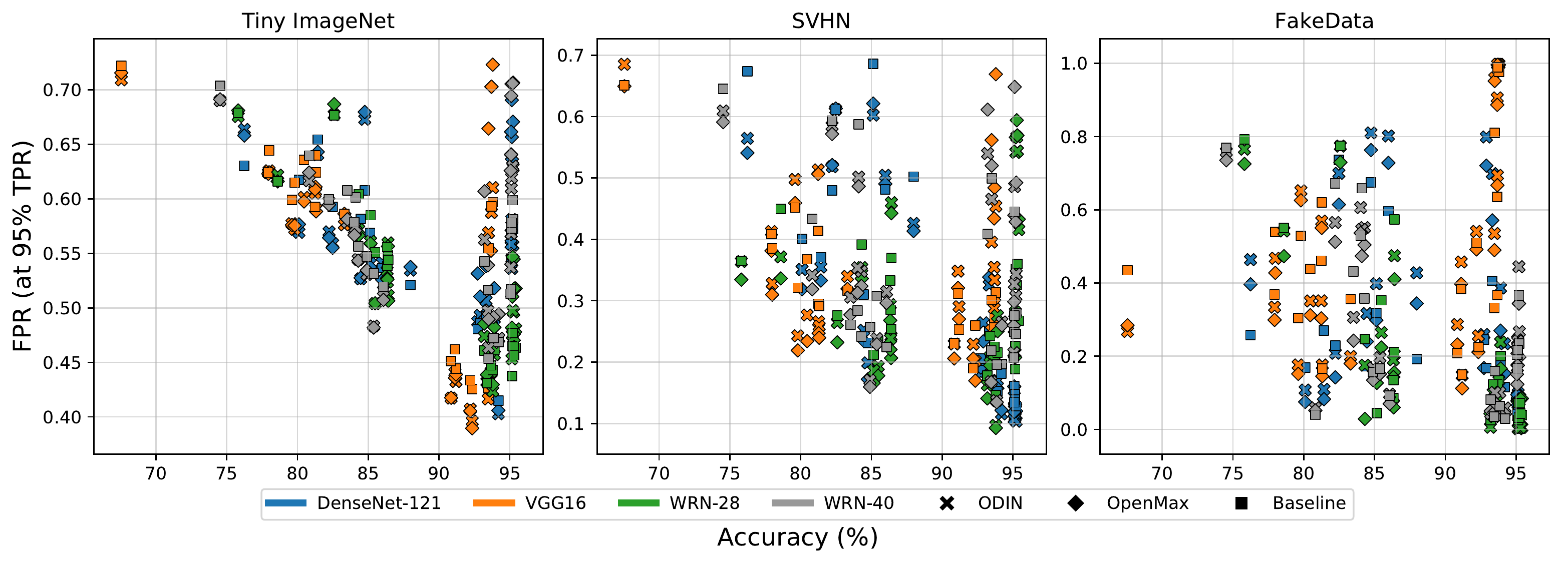}
\caption{The FPR at 95\% TPR (FPR95) metric plotted against the test set accuracy for each individual model, on the three out-of-distribution datasets. The coloring represents which model, whilst the marker represent which supervisor is used. i.e an orange square indicates one of the epochs from VGG16 tested with Baseline.}
\label{fig:FPR-vs-accuracy}
\end{figure*}

For coverage breakpoint on performance level, i.e., how restrictive the supervisors have to be to receive same overall accuracy as achieved on the test set, all models except for VGG16 perform similarly, independent on which supervisor is used. Most models range between 38\% and 43.5\% CBPL for Tiny ImageNet, which results in 13-24\% of the inlier samples being falsely rejected. It is noteworthy for DenseNet-121 to achieve $>50\%$ CBPL on SVHN and FakeData, which indicates the supervisor can exclude the majority outlier samples, as well as some inliers where the prediction would have been wrong. When changing the accuracy requirement to 90\%, the penalization on higher accuracy models has been lifted. This change elicit higher coverage, as the supervisor-model combination has more leeway. 

Overall, the DenseNet-121 model outperforms the rest regarding supervisor performance as well as coverage with and without diminishing requirements. Even though WRN-28 and WRN-40 receives slightly better accuracy on the test set, their performance in combination with a supervisor shows the problematic nature of tuning supervisors. As we will see in the following, these small variations in network weights can have large effects on the performance of the supervisor-model pair.  

\subsection{Results during training} 
\begin{figure*}[!ht]
\centering
\includegraphics[width=\textwidth]{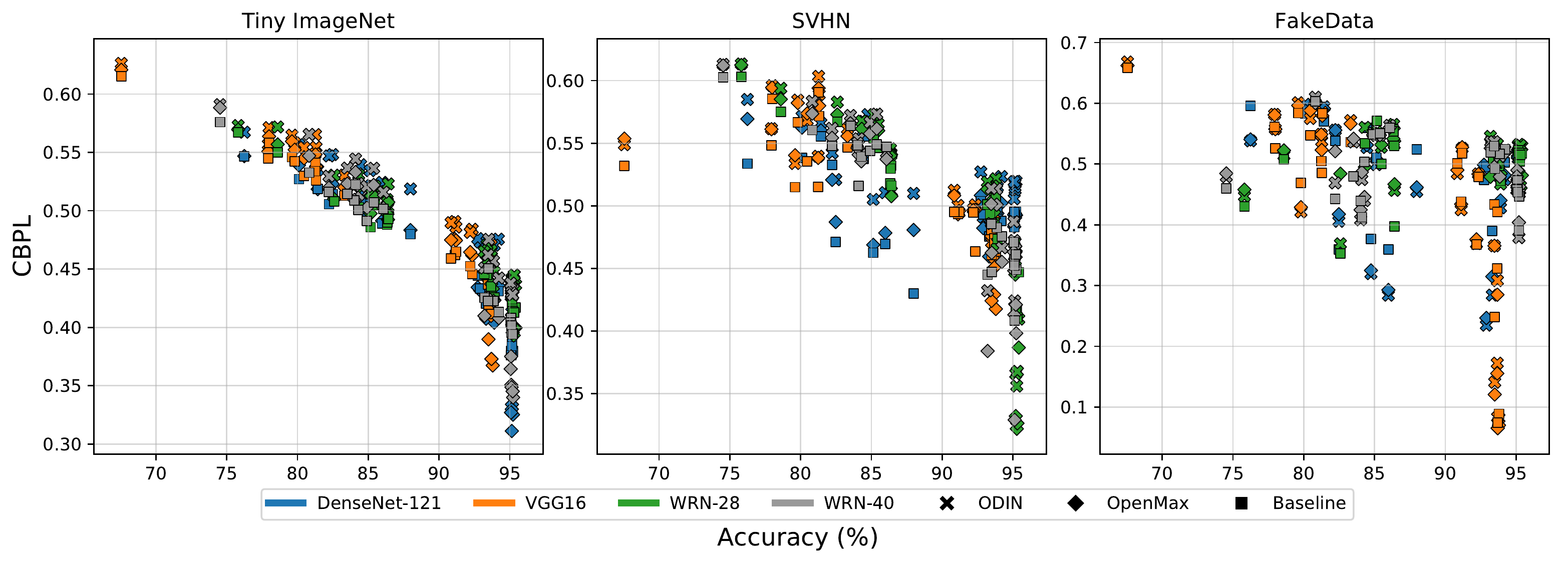}
\caption{The Coverage Breakpoint at Performance Level (CBPL) metric plotted against the test set accuracy for each individual model, on the three out-of-distribution datasets. The coloring represents which model, whilst the marker represent which supervisor is used. i.e an orange square indicates one of the epochs from VGG16 tested with Baseline.}
\label{fig:cbpl-vs-accuracy}
\end{figure*}

\begin{figure*}[!b]
\centering
\includegraphics[width=\textwidth]{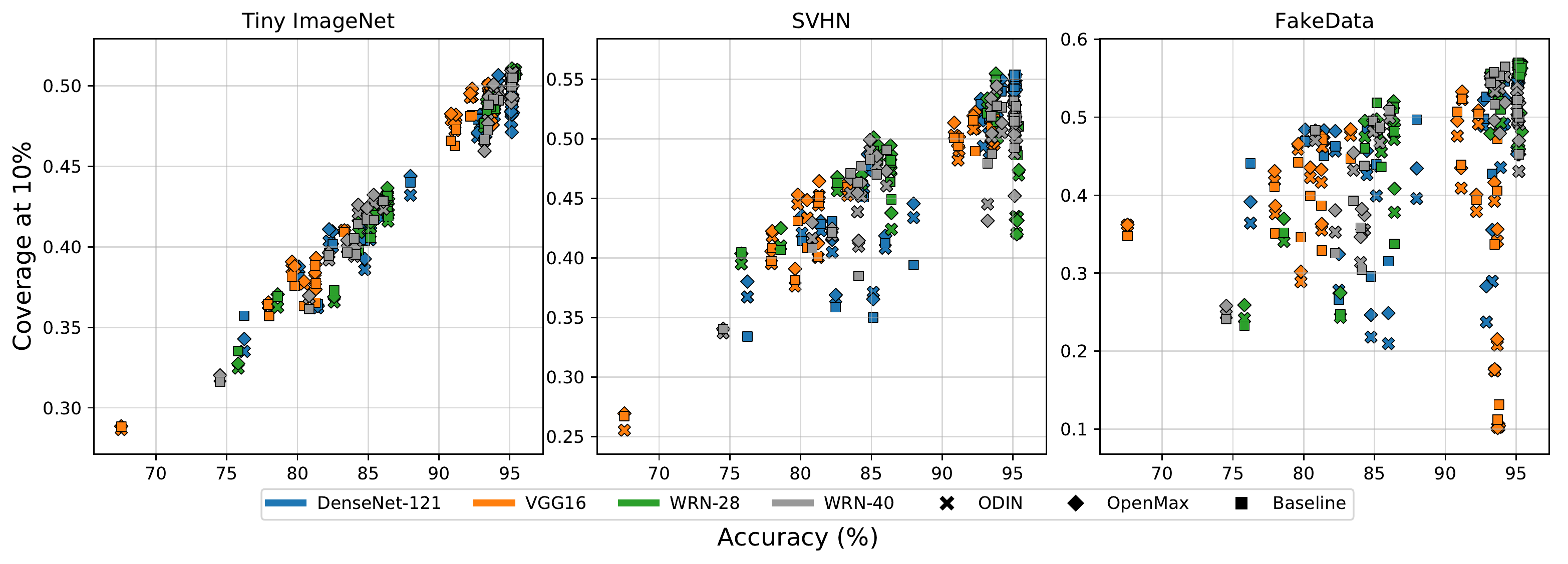}
\caption{The Coverage at 10\% error rate (Cov10) metric plotted against the test set accuracy for each individual model, on the three out-of-distribution datasets. The coloring represents which model, whilst the marker represent which supervisor is used; i.e., an orange square indicates one of the epochs from VGG16 tested with Baseline.}
\label{fig:cov10-vs-accuracy}
\end{figure*}

In Section \ref{sec:eval}, we explained why it is relevant to study the supervisor performance as the model is trained. Additionally, in this case, it is also relevant to study how the performance change as the model coverges and starts overfitting towards the training set.  In this section, our selected metrics are visualized in Fig.~\ref{fig:auroc-vs-accuracy}-\ref{fig:cov10-vs-accuracy}, respectively, and reviewed in further detail. All plots are designed the same: Color indicates which model has been used, and the marker indicates which supervisor has been used, i.e., an orange square indicates VGG16-Baseline. 

\textbf{AUROC vs Accuracy:} 
The AUROC scores can be seen in Fig.~\ref{fig:auroc-vs-accuracy}. As we presented in \cite{henriksson_automotive_2019}, we found a linear relationship between the model accuracy and supervisor performance. For the Tiny ImageNet dataset this pattern still exists, whereas SVHN and FakeData does not contain this clear behavior. While SVHN does not resemble a clear linear trend, it contains a positive lower boundary that increases as the model accuracy improves. For the FakeData, the results are instead scattered. It is unclear why this behavior exists, but could be due to the noise matrices are being randomly generated for experimentation, and not being kept static. Furthermore, for Tiny ImageNet and SVHN, the supervisor performance is independent from which model is selected. 

An interesting phenomenon occurs for small variations in model parameters. As have been expressed earlier, small subtle changes can have large impacts on the model output, as well as the performance of the supervisor. Looking at clusters that are present for both Tiny ImageNet and SVHN in the top right corners, this impact becomes evident. While a large amount of experiments are conducted on models with less than a tenth of a percentage difference, the AUROC score can vary as much as 10\% for Tiny ImageNet and up to 20\% in SVHN. This is worrisome, as there is no indication from the training setup, which yields a more separable end result. An initial explanation of this varying performance of OpenMax and ODIN can be explained by poor parameter settings, indicating the need a parameter search for each change of model weights. However, this does not explain why the behavior is also present for the Baseline supervisor that does not have any parameters.

\textbf{FPR at 95\% TPR vs Accuracy:} 
False positive rate, i.e., the amount of inlier data that has falsely been rejected, is a an effective metric to look at in comparison to True positive rate, i.e., the successful rejection of outlier samples. In this case, we investigate how many inlier samples have been rejected when the supervisor catches 95\% of the outliers. In combination with AUROC metric, this tells us several aspects about the model-supervisor combination: If the AUROC score is close to 1 and FPR95 is close to 0, it indicates the supervisor achieves a clear separation between the two distributions, whereas the AUROC decreases and FPR95 increases, there exists some overlapping of the distributions.   

The FPR95 vs.~accuracy plots can be seen in Fig.~\ref{fig:FPR-vs-accuracy}. At first, the two image datasets show a downward trend, i.e., as the model performance increases, the separation of in and out-of-distribution samples becomes more clear. All dataset experiments contain large variations of FPR95, see for example Tiny ImageNet and SVHN for models with $>95\%$ accuracy, where the FPR95 can vary between 0.45 up to 0.72. This is once again related to small variations in the model, which causes the robustness of the supervisor to drastically change. The variations seem to be related to selected models, for example in SVHN where DenseNet and WRN-28 performs well, but VGG16 and WRN-40 seem to fall short no matter which supervisor is used. For Tiny ImageNet, FPR95 never breaks below 0.37. This is similar to the AUROC case due to the datasets not being disjoint, and thus overlapping in distributions will be present. On the FakeData, VGG16 struggles more than the other models, especially for later epochs where the model is more accurate, as the behavior is not present as the model ranges between 85-92\% accuracy. This shows how the effect of overfitting harms the generalization of the model as well as the possibility for the supervisor to distinguish between distributions.

\textbf{Coverage breakpoints vs Accuracy:} 
While AUROC and FPR95 cover the overall performance of supervisor rejection rate, it is of interest to study what inlier samples get rejected first. By studying coverage, we can see if correctly classified samples or mis-classifications are the first ones to be rejected. 

Our previous work (cf.~\cite{henriksson2019performance}) used CBPL, similarly to this work, see Fig.~ \ref{fig:cbpl-vs-accuracy}. In the figure we see similar behavior as reported in our previous work, where the coverage is decreasing for more accurate models. This is to be expected as the restriction rate has to be stricter to achieve the high accuracy of the model. Studying Fig.~\ref{fig:cbpl-vs-accuracy} more in-depth, it can be seen that the average coverage of Tiny ImageNet is lower than to SVHN and FakeData, which shows how overlapping distributions increase difficulty in separation. 

Having 50\% coverage means the model is predicting on the same amount of samples as in the test set. Having more than 50\% coverage would indicate that the supervisor has managed to not only exclude outlier samples, but also some inlier samples that would have yielded a mis-classification. Thus the final result becomes a mix of excluding the majority of outlier samples, as well as a miniority of inlier samples with high probability of being mis-classified. We see this behavior for worse performing models in Tiny ImageNet as well as on several occasions in the SVHN and FakeData datasets.   

However, it seems non-intuitive to penalize well-performing models solely on the basis to reach same error rate as achieved during training. A more test-oriented approach would be to give the model a fixed target error rate, i.e., 10\%, which is higher than what all models achieved, and tune the threshold to maximize the coverage for the given error rate. We show this in Fig.~\ref{fig:cov10-vs-accuracy}. In this case, we can see how the coverage will increase with model performance and achieve models with more than 50\% coverage for the best performing models. Additionally, for models that have a larger error rate than the fixed target will now have to be more restrictive to reject outlier samples, as well as inliers that would be mis-classified to reach the target error rate.


\section{Discussion} \label{sec:disc}
As the results improve of DNNs on various tasks, the fact that the confidence level of out-of-distribution samples are high poses a problem to validity. Even with more sophisticated outlier detection tools, we have seen here that small adjustments in model weights can have large effects of the performance. Thus, the question arises what additional ways of measuring model generalization are needed in addition to the loss and accuracy metrics that are used during training. 

To deploy DNNs for safety critical applications, it is necessary to distinguish between \textit{known} states and \textit{unknown} states, i.e., it is better to know the functional limits,and detect scenarios that the system is not capable at handling than blindly believing it can handle anything. Additionally, since it is infeasible to record and train on all possible scenarios (cf.~\cite{borg_safely_2019}), instead it requires systematic ways of handling them to reduce the risk of the function to operate on scenarios it was not built for. 

To understand and be able to verify the performance of DNNs, additional experiments and methods are needed that explain the behavior of the model. This includes outlier detection, but also several additional topics such as uncertainty estimations and development procedures that argue for development processes, which aim to reduce the risk of mis-happenings with the model. As SOTIF was released to cover the quantitative requirements of functions, it still remains unclear how this should be set up.  

\section{Conclusions and Future Work} \label{sec:conc}
In this work we show how performance of supervisors change as the model performance varies. We test it after every $10^{th}$ epoch, and find that very small changes in the network, that does not yield any big difference in accuracy or loss, can affect the out-of-distribution detection significantly. Additionally, we show how the change of the coverage metric to a fixed error rate gives the model more leeway to reject samples not only from the outlier distribution but also from inlier samples where the risk of a miss-classification is high. 

There are many interesting aspects of continuing this research. Initially the field \textit{need} to start constructing safety argumentation that describes what and how methods that support the DNN shall be constructed and tested. SOTIF provides a starting point for these discussions, but industry needs to join to describe suitable development processes, and which measures are required to validate models that are learned from data. Additionally, when it comes to both model and supervisor exploration, it is desired to study what samples are rejected for the different outliers to see if an ensemble of models with a corresponding supervisor would yield a more stable result. 

\section*{Acknowledgments}
This work was carried out within the SMILE II project financed by Vinnova, FFI, Fordonsstrategisk forskning och innovation under the grant number: 2017-03066, and partially supported by the Wallenberg AI, Autonomous Systems and Software Program (WASP) funded by Knut and Alice Wallenberg Foundation.

\bibliographystyle{IEEEtran}
\bibliography{smile}

\end{document}